\documentclass[conference]{IEEEtran}
\IEEEoverridecommandlockouts
\usepackage{cite}
\usepackage{amsmath,amssymb,amsfonts}
\usepackage{algorithmic}
\usepackage{graphicx}
\usepackage{textcomp}
\usepackage{xcolor}

\def\BibTeX{{\rm B\kern-.05em{\sc i\kern-.025em b}\kern-.08em
    T\kern-.1667em\lower.7ex\hbox{E}\kern-.125emX}}
    
\begin{document}

\title{Visual question answering based evaluation metrics for text-to-image generation}

\author{\IEEEauthorblockN{1\textsuperscript{st} Mizuki Miyamoto}
\IEEEauthorblockA{
\textit{Hosei University}\\
Tokyo, Japan \\
mizuki.miyamoto.8b@stu.hosei.ac.jp}
\and
\IEEEauthorblockN{2\textsuperscript{nd} Ryugo Morita}
\IEEEauthorblockA{
\textit{Hosei University}\\
Tokyo, Japan \\
ryugo.morita.7f@stu.hosei.ac.jp}
\and
\IEEEauthorblockN{3\textsuperscript{rd} Jinjia Zhou}
\IEEEauthorblockA{
\textit{Hosei University}\\
Tokyo, Japan \\
zhou@hosei.ac.jp}
}

\maketitle

\begin{abstract}
Text-to-image generation and text-guided image manipulation have received considerable attention in the field of image generation tasks.
However, the mainstream evaluation methods for these tasks have difficulty in evaluating whether all the information from the input text is accurately reflected in the generated images, and they mainly focus on evaluating the overall alignment between the input text and the generated images.
This paper proposes new evaluation metrics that assess the alignment between input text and generated images for every individual object. 
Firstly, according to the input text, chatGPT is utilized to produce questions for the generated images. 
After that, we use Visual Question Answering(VQA) to measure the relevance of the generated images to the input text, which allows for a more detailed evaluation of the alignment compared to existing methods. 
In addition, we use Non-Reference Image Quality Assessment(NR-IQA) to evaluate not only the text-image alignment but also the quality of the generated images. 
Experimental results show that our proposed evaluation approach is the superior metric that can simultaneously assess finer text-image alignment and image quality while allowing for the adjustment of these ratios.
\end{abstract}


\begin{figure*}[t]
  \begin{center}
  \centering
  \includegraphics[width=\textwidth]{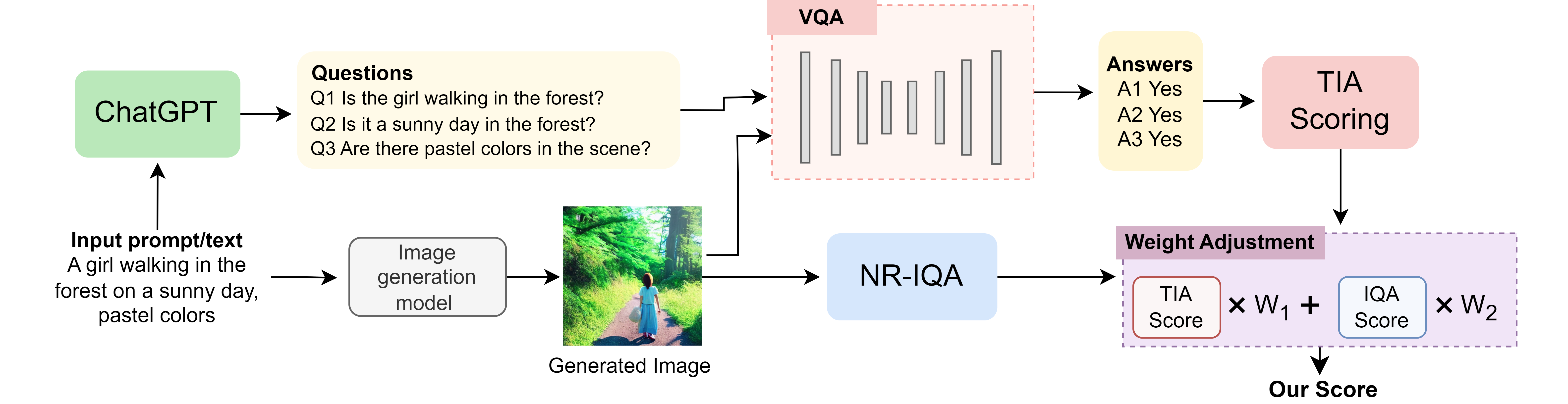}
  \caption{An overview of our method. chatGPT \cite{ouyang2022training} is utilized to generate questions for VQA according to the input text. VQA and NR-IQA are employed to evaluate the Text-Image Alignment (TIA) and image quality of the generated images. The TIA Scoring process involves quantifying the outputs from VQA. The final score is produced by adjusting the weighting between these two scores.}
  \label{fig:method}
  \end{center}
\end{figure*}

\section{Introduction}
In recent years, there has been significant progress in generative AI. In particular, image generation tasks, such as text-guided image manipulation \cite{li2020manigan,zhu2017unpaired} and text-to-image generation \cite{rombach2022high,saharia2022photorealistic}, have received considerable attention and experienced remarkable growth. These image generation tasks involve generating and manipulating images based on input text. These models are often trained on datasets such as COCO \cite{lin2014microsoft}, CUB \cite{WahCUB_200_2011}, DiffusionDB \cite{wang2022diffusiondb}, and comparisons of performance are conducted using existing evaluation metrics.

However, current automated evaluation metrics face several challenges. For example, Fréchet Inception Distance (FID) \cite{heusel2017gans} is a widely used metric in image generation that measures the realism of generated images by quantifying the distance between embeddings of real and generated images. However, FID \cite{heusel2017gans} doesn't consider the alignment with input text. Methods like image quality assessment can evaluate image quality but lack consideration for the text-image relationship like FID \cite{heusel2017gans} does. Another metric, CLIPScore \cite{hessel2021clipscore}, assesses the image-text relationship by measuring the cosine similarity between generated image features and text features. However, CLIPScore \cite{hessel2021clipscore} primarily emphasizes global feature similarity in the CLIP \cite{radford2021learning} space and often falls short in evaluating the precise alignment between generated images and given text.

Therefore, we propose a new automatic evaluation method for text-guided image manipulation and text-to-image generation. In this proposed method, we incorporate Visual Question Answering(VQA) model to assess whether the generated images are correctly conditioned on the input text. This enables a more detailed evaluation of the coherence between the input text and the generated images. Additionally, our method combines Non-Reference Image Quality Assessment(NR-IQA) to incorporate the evaluation of image quality along with the alignment between the generated images and the text. Furthermore, in our proposed method, the weighting of the text-image alignment score and the image quality score can be adjusted at the user's preference. This allows for a stronger influence of the preferred aspect's score on the final score, or the display of evaluations for either aspect alone.

Experimental results comparing our proposed method with CLIPScore \cite{hessel2021clipscore} and the existing NR-IQA method, MANIQA \cite{yang2022maniqa}, demonstrate that our approach can simultaneously reflect image quality assessment and more detailed text-image alignment in the scores. In addition, comparisons with the state-of-the-art evaluation method for text-to-image generation, ImageReward \cite{xu2023imagereward}, reveal the superiority of our method in that it achieves comparable accuracy in evaluating text-image alignment and image quality, and furthermore, the ratio of these evaluations can be arbitrarily adjusted.

To summarize, our contributions are twofold:
\begin{itemize}
    \item We introduce a novel automatic evaluation approach for text-guided image manipulation and text-to-image generation. This novel approach incorporates VQA and NR-IQA, allowing for the simultaneous evaluation of text-image alignment and image quality of generated images. Moreover, the weighting of these evaluations can be adjusted flexibly.
    \item Through comparisons with CLIPScore \cite{hessel2021clipscore}, MANIQA \cite{yang2022maniqa}, and ImageReward \cite{xu2023imagereward}, we have demonstrated the superiority of our method that offers simultaneous evaluation capability and adjustability.
\end{itemize} 

\section{Proposed Method}
As shown in Figure \ref{fig:method}, we employ a VQA model within the proposed network to assess the fidelity of the generated images to the input text or prompt. In the VQA task, an image and corresponding questions are provided as input, and the model generates an answer to the questions. By utilizing this task, we can directly inquire whether the information contained in the text is reflected in the generated image. In addition, by incorporating NR-IQA, we enable the simultaneous evaluation of both the text-image alignment and the quality of the generated image.

\subsection{Question generating}
We utilize a VQA model to evaluate the match between the generated image and the input text or prompt. The questions used in VQA are automatically generated by chatGPT \cite{ouyang2022training} based on the input text or prompt used during the image generation process. Using chatGPT \cite{ouyang2022training}, we can automatically generate question sentences from the input text or prompt closer to natural language expressions. When generating questions with chatGPT \cite{ouyang2022training}, we have specified several conditions. Firstly, the questions are generated as simple sentences of about seven words. The aim is to increase the accuracy of the VQA model. Additionally, we have specified that the generated question sentences should all have the answer "Yes", enabling the calculation of accuracy by simply counting the number of "yes" responses. The number of generated questions is determined based on the number of words in the input text or prompt. Specifically, if the input text or prompt consists of 1 or 2 words, one question is generated. For input text or prompt with more than 2 words, an additional question is generated for every 6-word increment. This specification is based on experimental results and is determined to be the most appropriate number of questions that can be generated without excess or deficiency. 

\subsection{Evaluation method of text-to-image alignment}
We incorporate VQA to evaluate whether the images are generated according to the instructions given by the input text or prompt. Based on several questions generated from the text, we directly verify through a VQA model whether all the information from the text is correctly represented in the generated images. We employ BEIT-3 \cite{wang2022image} as the VQA model, which utilizes multiway transformers as a fusion encoder to extract and fusion features from both the images and the text. The classifier layer predicts the answers to the questions, which serve as the final outputs from the VQA model. Then Text-Image Alignment Scoring is performed, where the obtained answers are aggregated, and the "Yes" answers are counted to calculate the accuracy rate. This accuracy rate serves as a score indicating the alignment between the generated images and the input text or prompt.

\subsection{Image quality assessment}
We evaluate not only the relevance between images and text but also the image quality of the generated images using NR-IQA. NR-IQA aims to automatically assess the perceptual quality of images(e.g., blur, JPEG, noise) without the reference images. We adopt MANIQA \cite{yang2022maniqa} as the NR-IQA model . This model performs feature extraction of input images using Vision Transformers \cite{dosovitskiy2020image} and then applies attention mechanisms \cite{vaswani2017attention} to enhance interactions between different image regions. After this operation, the final score prediction is performed. Higher scores indicate better perceptual quality of the images.

\begin{figure*}[t]
  \begin{center}
  \centering
  \includegraphics[width=\textwidth]{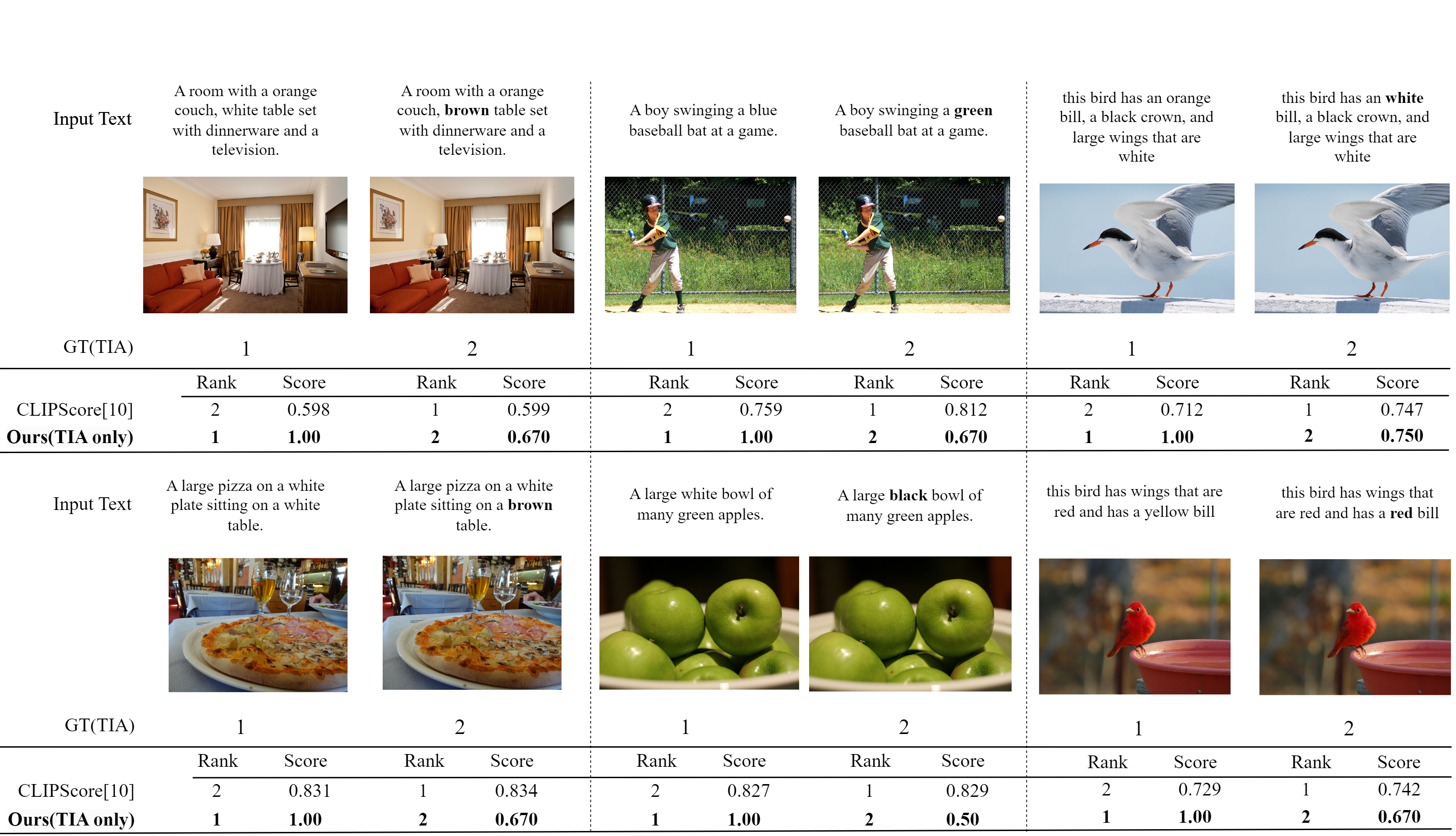}
  \caption{Comparison of text-image alignment scores with CLIPScore \cite{hessel2021clipscore}. There are distinct texts provided for identical two images. One text aligns with the actual content of the image, while the other contains expressions inconsistent with the image content. The bold in the input text represents words that do not align with the image representation. GT(TIA) indicates the rank of the Text-Image Alignment.}
  \label{fig:TIA}
  \end{center}
\end{figure*}

\begin{figure*}[t]
  \begin{center}
  \centering
  \includegraphics[width=\textwidth]{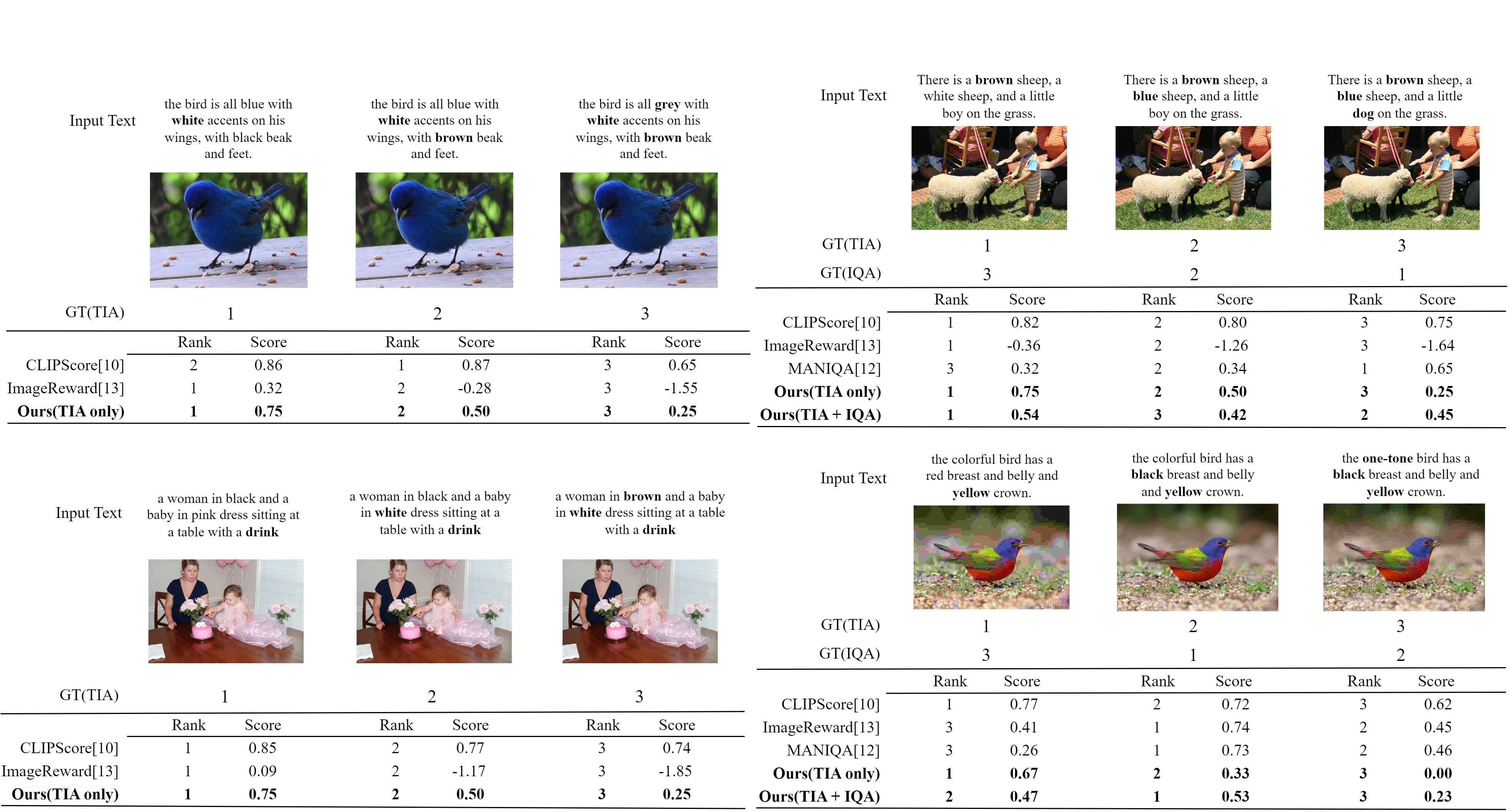}
  \caption{\textbf{left: }Comparison of text-image alignment scores with CLIPScore \cite{hessel2021clipscore} and ImageReward \cite{xu2023imagereward}. There are distinct texts provided for identical three images. The bold in the input text represents mismatched words with the image representation. GT(TIA) indicates the rank of the Text-Image Alignment.  \textbf{right: }Comparison of text-image alignment and image quality scores with CLIPScore \cite{hessel2021clipscore}, ImageReward \cite{xu2023imagereward}, and MANIQA \cite{yang2022maniqa}. Applying degradation operations to three images with different JPEG compression rates. GT(IQA) represents the rank of Image Quality Assessment. Ours(TIA + IQA) score is the combination of TIAscore and IQAscore with equal weighting.
  }
  \label{fig:IQA}
  \end{center}
\end{figure*}

\subsection{Adjustable final score}
The final output score is a weighted sum of the score representing the Text-Image Alignment(TIA) and the Image Quality Assessment(IQA) of the generated images. These scores are individually weighted and then combined. The user can adjust the weights based on their preferences, such as giving more importance to one score over the other. The calculation of the final score is formulated as:
\begin{equation}
   S_{out} = S_{tia} * W_1  +  S_{iqa} * W_2
 \end{equation}

Where 
\begin{math}
   S_{out}
\end{math} 
represents our final output score, and 
\begin{math}
   S_{tia}, S_{iqa}, W_1,
\end{math}
and
\begin{math}
   W_2
\end{math}
denote the TIAscore, IQAscore, and the weights for adjusting the ratio, respectively. Based on experimental results, we set the default values for 
\begin{math}
   W_1
\end{math}
and
\begin{math}
   W_2
\end{math}
to be equal, 
\begin{math}
   W_1=W_2=0.5
\end{math}
, to give equal importance to both TIAscore and IQAscore.

\section{experiment}
Our evaluation method is evaluated on the CUB \cite{WahCUB_200_2011} and the COCO \cite{lin2014microsoft} datasets, comparing with existing evaluation methods for text-to-image generation, CLIPScore \cite{hessel2021clipscore} and ImageReward \cite{xu2023imagereward}, and with an existing NR-IQA method, MANIQA \cite{yang2022maniqa}. To compare the image quality assessment, we applied operations such as blur, Gaussian noise, and JPEG compression to some test images.  As input text, we utilized the provided CUB \cite{WahCUB_200_2011} and COCO \cite{lin2014microsoft} annotations. To evaluate the TIAscores, we modified some words in the text.

\subsection{Comparison with CLIPScore}
Figure \ref{fig:TIA} shows the comparison of our method with CLIPScore \cite{hessel2021clipscore} in terms of text-image alignment evaluation. In this figure, two types of text inputs are provided for a single image: one text input that matches the image representation and another text input that contains a single mismatched word. This is designed to account for situations where the image is generated exactly as described in the input text, and situations where the generated image only partially reflects the input text. 

From this figure, it can be observed that CLIPScore \cite{hessel2021clipscore} yields higher scores to pairs with lower text-image alignment. This indicates that CLIPScore \cite{hessel2021clipscore} fails to accurately evaluate the fine-grained alignment between the text and image representations. On the other hand, our method consistently demonstrate that the scores increase with increasing text-image alignment across all rows, affirming their ability to finely evaluate the alignment between text and image.

\subsection{Comparison of Text-Image Alignment evaluations}
The left side of Figure \ref{fig:IQA} shows the comparison of our method with CLIPScore \cite{hessel2021clipscore} and ImageReward \cite{xu2023imagereward} in terms of text-image alignment evaluation. In each row, the leftmost column calculates the scores based on the image and the text that accurately describes the information in the image. This assumes that the image is generated faithfully according to the input text. In the other three columns, the scores are calculated based on the image and the text containing words that do not match the image. This is done to account for cases where images are generated that are partially or completely unrelated to the input text. 

In the top row, it can be observed that CLIPScore \cite{hessel2021clipscore} yields higher scores for the lower TIA columns. This indicates that CLIPScore \cite{hessel2021clipscore} fails to accurately evaluate the fine-grained alignment between the text and image representations. On the other hand, our method and ImageReward \cite{xu2023imagereward} consistently show that the scores decrease as the text-image alignment decreases in all rows, indicating that they are capable of finely evaluating the alignment between text and image.

\subsection{Comparison of TIA and Image Quality evaluations}
The right side of Figure \ref{fig:IQA} shows the comparison of text-image alignment and image quality evaluations between our method and CLIPScore \cite{hessel2021clipscore}, ImageReward \cite{xu2023imagereward}, MANIQA \cite{yang2022maniqa}. In this figure, we calculate scores between various TIA levels of text and images with different qualities to perform a more practical evaluation. 

In this figure, it can be observed that CLIPScore \cite{hessel2021clipscore} is accurately evaluating TIA, but it lacks the ability to assess image quality, thus it cannot evaluate degradation caused by noise, blur, or JPEG. MANIQA \cite{yang2022maniqa} shows decreasing scores as the noise level increases, providing an accurate evaluation of image quality for all images. However, it is incapable of assessing the match between text and image. In contrast, our method and ImageReward \cite{xu2023imagereward} exhibits decreasing scores as the agreement between text and image decreases and as the image quality deteriorates. This indicates that our method and ImageReward \cite{xu2023imagereward} can finely evaluate the text-image relevance, while also accurately assessing image quality. 

However, ImageReward \cite{xu2023imagereward} cannot separate the scores for TIA and IQA. In the middle row, the ImageReward's score increases even as the image quality degrades, making it difficult to verify if the image quality evaluation is correctly reflected in the score. On the other hand, our method allows separate computation of alignment and image quality scores, enabling simultaneous verification of both alignment and image quality evaluations in numerical form even in such situations. Additionally, in our method, the emphasis on TIA or IQA can be adjusted at the user's discretion. For instance, in the right side of Figure \ref{fig:IQA}, we have combined the TIAscores and IQAscores with equal weighting. However, it is also possible to increase the ratio of either one to more strongly reflect in the final score.

\section{Conclusion}
In this study, we propose a novel evaluation method for text-guided image manipulation and text-to-image generation. Our method incorporates the VQA to provide a more detailed assessment of the text-image agreement. Additionally, we employ NR-IQA to evaluate the quality of the generated images. Our evaluation score allows for the adjustment of the TIAscore and IQAscore ratio according to user preferences. The experimental results demonstrate the superiority of our approach, which can simultaneously evaluate text-image alignment and image quality, while allowing for the adjustment of their ratios.

\bibliographystyle {IEEEtran}
\bibliography{sample-base}

\begin{thebibliography}{10}
\providecommand{\url}[1]{#1}
\csname url@samestyle\endcsname
\providecommand{\newblock}{\relax}
\providecommand{\bibinfo}[2]{#2}
\providecommand{\BIBentrySTDinterwordspacing}{\spaceskip=0pt\relax}
\providecommand{\BIBentryALTinterwordstretchfactor}{4}
\providecommand{\BIBentryALTinterwordspacing}{\spaceskip=\fontdimen2\font plus
\BIBentryALTinterwordstretchfactor\fontdimen3\font minus \fontdimen4\font\relax}
\providecommand{\BIBforeignlanguage}[2]{{%
\expandafter\ifx\csname l@#1\endcsname\relax
\typeout{** WARNING: IEEEtran.bst: No hyphenation pattern has been}%
\typeout{** loaded for the language `#1'. Using the pattern for}%
\typeout{** the default language instead.}%
\else
\language=\csname l@#1\endcsname
\fi
#2}}
\providecommand{\BIBdecl}{\relax}
\BIBdecl

\bibitem{ouyang2022training}
L.~Ouyang, J.~Wu, X.~Jiang, D.~Almeida, C.~Wainwright, P.~Mishkin, C.~Zhang, S.~Agarwal, K.~Slama, A.~Ray \emph{et~al.}, ``Training language models to follow instructions with human feedback,'' \emph{Advances in Neural Information Processing Systems}, vol.~35, pp. 27\,730--27\,744, 2022.

\bibitem{li2020manigan}
B.~Li, X.~Qi, T.~Lukasiewicz, and P.~H. Torr, ``Manigan: Text-guided image manipulation,'' in \emph{Proceedings of the IEEE/CVF Conference on Computer Vision and Pattern Recognition}, 2020, pp. 7880--7889.

\bibitem{zhu2017unpaired}
J.-Y. Zhu, T.~Park, P.~Isola, and A.~A. Efros, ``Unpaired image-to-image translation using cycle-consistent adversarial networks,'' in \emph{Proceedings of the IEEE international conference on computer vision}, 2017, pp. 2223--2232.

\bibitem{rombach2022high}
R.~Rombach, A.~Blattmann, D.~Lorenz, P.~Esser, and B.~Ommer, ``High-resolution image synthesis with latent diffusion models,'' in \emph{Proceedings of the IEEE/CVF conference on computer vision and pattern recognition}, 2022, pp. 10\,684--10\,695.

\bibitem{saharia2022photorealistic}
C.~Saharia, W.~Chan, S.~Saxena, L.~Li, J.~Whang, E.~L. Denton, K.~Ghasemipour, R.~Gontijo~Lopes, B.~Karagol~Ayan, T.~Salimans \emph{et~al.}, ``Photorealistic text-to-image diffusion models with deep language understanding,'' \emph{Advances in Neural Information Processing Systems}, vol.~35, pp. 36\,479--36\,494, 2022.

\bibitem{lin2014microsoft}
T.-Y. Lin, M.~Maire, S.~Belongie, J.~Hays, P.~Perona, D.~Ramanan, P.~Doll{\'a}r, and C.~L. Zitnick, ``Microsoft coco: Common objects in context,'' in \emph{Computer Vision--ECCV 2014: 13th European Conference, Zurich, Switzerland, September 6-12, 2014, Proceedings, Part V 13}.\hskip 1em plus 0.5em minus 0.4em\relax Springer, 2014, pp. 740--755.

\bibitem{WahCUB_200_2011}
C.~Wah, S.~Branson, P.~Welinder, P.~Perona, and S.~Belongie, ``The caltech-ucsd birds-200-2011 dataset,'' California Institute of Technology, Tech. Rep. CNS-TR-2011-001, 2011.

\bibitem{wang2022diffusiondb}
Z.~J. Wang, E.~Montoya, D.~Munechika, H.~Yang, B.~Hoover, and D.~H. Chau, ``Diffusiondb: A large-scale prompt gallery dataset for text-to-image generative models,'' \emph{arXiv preprint arXiv:2210.14896}, 2022.

\bibitem{heusel2017gans}
M.~Heusel, H.~Ramsauer, T.~Unterthiner, B.~Nessler, and S.~Hochreiter, ``Gans trained by a two time-scale update rule converge to a local nash equilibrium,'' \emph{Advances in neural information processing systems}, vol.~30, 2017.

\bibitem{hessel2021clipscore}
J.~Hessel, A.~Holtzman, M.~Forbes, R.~L. Bras, and Y.~Choi, ``Clipscore: A reference-free evaluation metric for image captioning,'' \emph{arXiv preprint arXiv:2104.08718}, 2021.

\bibitem{radford2021learning}
A.~Radford, J.~W. Kim, C.~Hallacy, A.~Ramesh, G.~Goh, S.~Agarwal, G.~Sastry, A.~Askell, P.~Mishkin, J.~Clark \emph{et~al.}, ``Learning transferable visual models from natural language supervision,'' in \emph{International conference on machine learning}.\hskip 1em plus 0.5em minus 0.4em\relax PMLR, 2021, pp. 8748--8763.

\bibitem{yang2022maniqa}
S.~Yang, T.~Wu, S.~Shi, S.~Lao, Y.~Gong, M.~Cao, J.~Wang, and Y.~Yang, ``Maniqa: Multi-dimension attention network for no-reference image quality assessment,'' in \emph{Proceedings of the IEEE/CVF Conference on Computer Vision and Pattern Recognition}, 2022, pp. 1191--1200.

\bibitem{xu2023imagereward}
J.~Xu, X.~Liu, Y.~Wu, Y.~Tong, Q.~Li, M.~Ding, J.~Tang, and Y.~Dong, ``Imagereward: Learning and evaluating human preferences for text-to-image generation,'' \emph{arXiv preprint arXiv:2304.05977}, 2023.

\bibitem{wang2022image}
W.~Wang, H.~Bao, L.~Dong, J.~Bjorck, Z.~Peng, Q.~Liu, K.~Aggarwal, O.~K. Mohammed, S.~Singhal, S.~Som \emph{et~al.}, ``Image as a foreign language: Beit pretraining for all vision and vision-language tasks,'' \emph{arXiv preprint arXiv:2208.10442}, 2022.

\bibitem{dosovitskiy2020image}
A.~Dosovitskiy, L.~Beyer, A.~Kolesnikov, D.~Weissenborn, X.~Zhai, T.~Unterthiner, M.~Dehghani, M.~Minderer, G.~Heigold, S.~Gelly \emph{et~al.}, ``An image is worth 16x16 words: Transformers for image recognition at scale,'' \emph{arXiv preprint arXiv:2010.11929}, 2020.

\bibitem{vaswani2017attention}
A.~Vaswani, N.~Shazeer, N.~Parmar, J.~Uszkoreit, L.~Jones, A.~N. Gomez, {\L}.~Kaiser, and I.~Polosukhin, ``Attention is all you need,'' \emph{Advances in neural information processing systems}, vol.~30, 2017.

\end{thebibliography}

\end{document}